\pdfoutput=1

\documentclass[twocolumn]{bmcart}

\usepackage[utf8]{inputenc} 
\usepackage{amssymb}
\usepackage{amsmath}
\usepackage{hyperref}
\usepackage{graphicx}
\usepackage{xcolor}
\usepackage{multirow}
\usepackage{array}

\usepackage{verbatim}

\startlocaldefs
\endlocaldefs

\begin{document}

\begin{frontmatter}

\begin{fmbox}
\dochead{Research}


\title{An Explainable CNN Approach for Medical Codes Prediction from Clinical Text}


\author[
   addressref={aff1},                   
   email={316615197@qq.com}   
]{\inits{Hu S.Y.}\fnm{ShuYuan} \snm{Hu}}
\author[
   addressref={aff2},
   corref={aff2}, 
   email={fteng@swjtu.edu.cn}
]{\inits{Teng F.}\fnm{Fei} \snm{Teng}}


\address[id=aff1]{
  \orgname{Leeds joint School, SWJTU}, 
  \city{Cheng Du},                              
  \cny{China}                                    
}
\address[id=aff2]{
  \orgname{The school of information science and technology, SWJTU}, 
  \city{Cheng Du},                              
  \cny{China}                                    
}


\begin{artnotes}
\end{artnotes}



 \begin{abstractbox}

\begin{abstract} 
\parttitle{Background} 
Clinical notes are unstructured text documents generated by clinicians during patient encounters, generally are annotated with ICD codes, which give formatted information about the diagnosis and treatment. ICD code has shown its potentials in many fields, but manual coding is labor-intensive and error-prone, lead to researches of automatic coding. Two specific challenges of this task are 1) given an annotated clinical notes, the reasons behind specific diagnoses and treatments are lost; 2) explainability is important for practical automatic coding method, the method should not only explain its prediction output but also have explainable internal mechanics. This study aims to develop an explainable CNN approach to address these two challenges.

\parttitle{Method} 
We develop CNN-based methods for automatic ICD coding based on clinical text from the intensive care unit (ICU) stays. We come up with the Shallow and Wide Attention convolutional Mechanism (SWAM), which allows our model to learn local and low-level features for each label. The key idea behind our model design is to look for the presence of informative snippets in the clinical text that correlated with each code, and we infer that there exists a correspondence between “informative snippet” and convolution filter.

\parttitle{Results}
We evaluate our approach on MIMIC-III, an open-access dataset of ICU medical records. Our approach substantially outperforms previous results on top-50 medical code prediction on MIMIC-III dataset. We attribute this improvement to SWAM, by which the wide architecture gives the model ability to more extensively learn the unique features of different codes, and we prove it by ablation experiment. Besides, we perform manual analysis of the performance imbalance between different codes, and preliminary conclude the characteristics that determine the difficulty of learning specific codes.

\parttitle{Conclusions}
We present an explainable CNN approach for multi-label document classification, which employs a wide convolution layer to learn local and low-level features for each label,  yields strong improvements over previous metrics on the ICD-9 code prediction task while providing satisfactory explanations for its internal mechanics.
\end{abstract}




\end{abstractbox}

\end{fmbox}

\end{frontmatter}



\section*{Background}
Clinical notes are written by clinicians during patient encounters, they are usually unstructured text narratives and accompanied by a set of metadata codes from the International Classification of Diseases (ICD), which present a standardized way of indicating diagnoses and procedures that were performed during the encounter. There is much research that demonstrates the practical application with ICD codes \cite{crawford2010,ranganath2015survival,avati2018}. For example in work by Choi et al. \cite{choi2016}, they proposed the Doctor AI system based on the presence of ICD codes to predict future patient states from learning patient representation from a large dataset of patient records. 

But manual coding is time-consuming and error-prone, so much research on automatic coding has been done in the past three decades, some recent works are Zhang et al. \cite{zhang2017}; Kavuluru et al. \cite{kavuluru2015} and Avati et al. \cite{avati2018}. And automatic coding is considered a multi-label classification task. Two domain-specific challenges are facing this task. 
First, a reasonable guess is that for a certain code prediction task, most of the text is not informative, only a few snippets are related to the code. However given the annotated text, the connections between code and its corresponding informative snippets are lost, in other words, the model has to learn the reasons behind specific diagnoses and treatments; second, interpretability is a crucial obstacle for practical automatic coding in both perspective of inferring and internal mechanics, the method is supposed to explain its prediction as well as have an explainable internal mechanics.

To address these two specific challenges together, in this paper, we develop CNN-based methods for automatic ICD code assignment based on text discharge summaries from intensive care unit (ICU) stays, we come up with Shallow and Wide Attention convolutional Mechanism (SWAM), which allows our model to learn local and low-level features for each label. Our model design is motivated by the way human clinicians manual label the clinical notes, which is to look for informative snippets that are relevant to each code. We consider the “informative snippets” as local and low-level features. SWAM address the two mentioned challenges in automatic coding: first, by transferring the base representation (i.e. clinical notes in the word-embedding form) to the convolution representation which represents the presence of “informative snippets”, the model could filter out the irrelevant information in the text, and through the attention mechanism the model could learn the 
correlation between “informative snippets” and labels; second, SWAM gives “informative snippets” extracted from clinical notes as explanations of its prediction result, and provides a new perspective for understanding the internal mechanics of the machine learning method.

We evaluate our approach on the MIMIC-III dataset \cite{johnson2016}, an open dataset of ICU medical records. With the Shallow and Wide Attention CNN mechanism, the model can learn non-generic features associated with specific labels that are not informative for other labels, which the narrow one are failed to learn. With the performance improvement gained from these specific labels, our approach outperforms previous results on medical code prediction on MIMIC-III dataset.

\subsection*{Related work}
\subsubsection*{Automatic ICD coding}

ICD coding has been a long-established task in the medical informatics community for decades, from the perspective of data, the current approaches of this task can be divided into two factions: much recent research focuses on unstructured text data \cite{subotin2014,kavuluru2015}, while the other incorporates structured data as well \cite{scheurwegs2017selecting}. We develop our methods on unstructured text data from the MIMIC3. Also, from the perspective of the code set, many approaches \cite{wang2016,prakash2017condensed} evaluate on a subset of the full ICD label space, while there are also methods \cite{mullenbach2018expICD} developed on the full code set. We develop our methods on the top-50 code set because the advantage of SWAM is learning specific features associated with specific labels that are not generic feature for other labels, so instead of carrying out a surprisingly large network to learn all non-generic features on the full code set, using ensemble method to cover the whole code set is preferred, which is discussed in later part.

A tendency in recent years is developing Neural Network-based methods for this task. Shi et al. \cite{shi2017towards}
applied attentional LSTMs to form a soft matching between sentence representations from discharge summaries and the top 50 codes. Prakash et al. \cite{prakash2017condensed} generated predictions of the top 50 codes by memory networks built from discharge summaries and Wikipedia. Mullenbach et al. \cite{mullenbach2018expICD} applied a per-label mechanism to extract the most important snippet for each code from discharge summaries. SWAM is compared with the published result from all these papers, and it achieves state-of-the-art results across many indicators. We attribute these improvements to the ability to learn non-generic features associated with specific labels that are not informative for other labels, which bring significant performance improvements on these specific labels.

\subsubsection*{Attentional convolution for NLP and explainable text classification}

Combing convolution with attention has been proved is efficient in different tasks among NLP \cite{allamanis2016convolutional,yin2016abcnn,santos2016attentive,yin2017attentive,bahdanau2014neural}. Yang et al. \cite{yang2016hierarchical} and  Mullenbach et al. \cite{mullenbach2018expICD} utilize attentional convolution to select the most relevant parts of the clinical text of each code. We refer to the per-label attention mechanism from those of Mullenbach et al. \cite{mullenbach2018expICD}, in which per-label parameter vectors are used to compute attention over specific locations in the text. Our work differs in that SWAM establishes the correspondence between the “informative snippet” and convolution filter, which makes the network a wider one comparing to its of Mullenbach et al. \cite{mullenbach2018expICD} and is better tuned to our goal of learning low-level feature, a.k.a. informative snippet with explainable internal mechanics. 

\begin{figure*}[h]

 \center

  \includegraphics[width=0.95\textwidth]{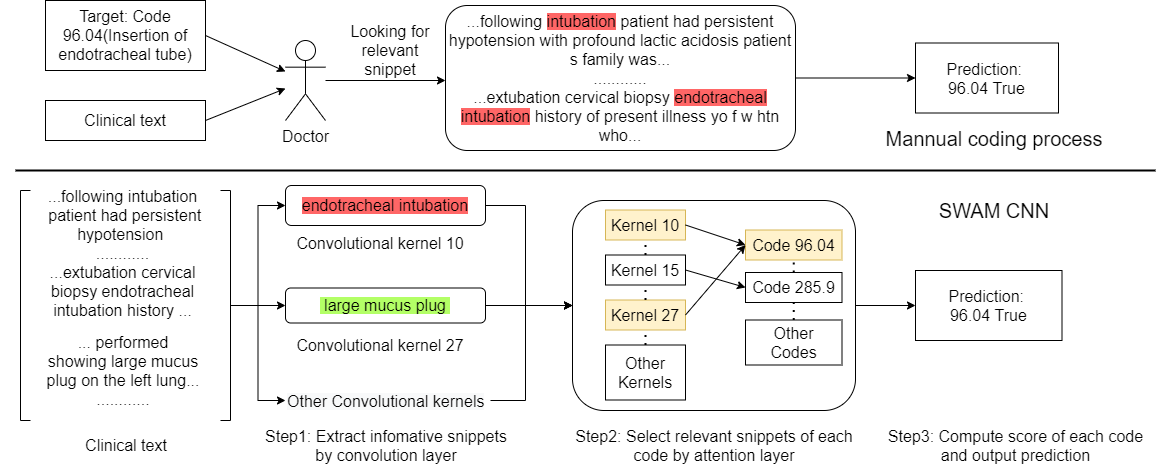}

  \caption{Internal mechanism of SWAM.}

  \label{illustrate}

\end{figure*}
Attentional Convolution has also been applied to make explainable text classification. Some prior works like Rushet al. \cite{rush2015neural} and Rocktäschel et al. \cite{rocktaschel2015reasoning} employ attention to highlight salient features of the text. The per-label attention mechanism \cite{mullenbach2018expICD} we referred extract snippet from the text as automatically generated explanation of the prediction in the same medical codes prediction task, and the informativeness of explanations are rated by a physician. Their work illustrates that the neural network work in an explainable way for this task: the model will try to find parts of the text that are most relevant to each code. Our work differs in that instead of making the model explainable by explaining its prediction, we take a further step forward to make the internal mechanics of the method explainable by opening the black box of the neural network to establish the correspondence between the “informative snippet” and convolution filter. We also bring out a preliminary analysis of the imbalance performance between the labels, provide a rational explanation of why the model performs terribly on certain codes.

\subsubsection*{Neural network architecture design for text classification work}

Hoa et al. \cite{le2017convolutional} compared the deep CNN and shallow CNN under text classification task, a practical rule is summarized that deep models do not seem to bring a significant advantage over shallow networks for text classification, another observation they made is that a global max-pooling \cite{collobert2008unified}, which retrieves the most influential feature could already be good enough for the text classification task. The authors believe one possible reason may be related to these facts that images are represented as real and dense values, as opposed to the discrete, artificial, and sparse representation of text. Their work indicates that local and low-level features extracted by shallow CNN work well for text classification tasks and inspires us to explore the underlying correspondence between local and low-level features and snippets in the text.

Gong and Ji \cite{gong2018does} find that in CNN for the text classification task, the convolution filters have learned division of labor. More than half of the kernels have a preference for one specific label. Their work inspires us to associate the width of the network with the learning of features of specific labels that are not generic for other labels.

\begin{table}[t]
  \caption{Table of Notations}
  \label{notation}
  \centering
  \begin{tabular}{lp{4cm}p{5cm}}
  \hline
  \textbf{Notation} &\textbf{Description} \\
  \hline
  $\mathcal{L}$     &   The set of ICD-9 codes.\\
  
  $y_{i,\ell}\in\ {0,\ 1}$ & The true value of the label task for instance i and $\ell\in\mathcal{L}$, 1 indicates the label is true for instance i. \\
  
  $d_e$ & The size of the input embedding \\
  
  $d_c$ & The size of the convolution output, a.k.a. the number of convolution filters \\
  
  $\mathbf{X}=[\mathbf{x}_1,\mathbf{x}_2,\ldots,\mathbf{x}_N]$ & The matrix of a document instance, where $\mathbf{N}$ is the length of the document and $\mathbf{x}_i$ is the vector representation of the word. \\
  
  $\mathbf{W}_c\in\mathbb{R}^{k\times d_e\times d_c}$ & Convolution filters, where k is the width of filter window. \\
  
  $\mathbf{H}\in\mathbb{R}^{d_c\times N}$ & Convolutional representation of the document. \\
  
  $\ast$ & Convolution operator. \\
  
  $g$ & An element-wise nonlinear transformation. \\
  
  $\mathbf{b}_c\in\mathbb{R}^{d_c}$ & The bias in convolutional operation. \\
  
  $\mathbf{u}_\ell\in\mathbb{R}^{d_c}$ & Attention parameter vector  for label. $\ell$ \\ 

  $\boldsymbol\alpha_\ell\in\mathbb{R}^N$ & Attention result vector for label. $\ell$ \\
  
  $b_\ell$ & Scalar offset in linear layer for label. $\ell$  \\
  
  $\boldsymbol\beta_\ell\in\mathbb{R}^{d_c}$ & Vector of prediction weights.  \\
  
  $\sigma$ & Sigmoid function.  \\
  
  $\operatorname { SoftMax }()$ & $\operatorname { SoftMax }(\mathbf{x})=\frac{\operatorname{\exp}(\mathbf{x})}{\sum_{i}{\operatorname{\exp}(x_i)}}$ , where $\operatorname{\exp}(\mathbf{x})$ is the element-wise exponentiation of the vector $\mathbf{x}$.  \\
  \hline
  \end{tabular}
\end{table}

\section*{Methods}
In this paper, we use notations shown in Table \ref{notation}.

We present SWAM, a CNN-based method for automatic ICD coding from the clinical text, which provides a good explanation of its internal mechanics.

SWAM is motivated by the way human clinicians manual label the clinical notes, to help the reader understand the method, firstly here is a brief introduction of the way human clinicians manual label the clinical notes. Normally, human clinicians will look for informative snippets that are relevant to each code. For example, as shown in Figure \ref{illustrate}, given code 96.04 in the figure, a human clinician will look for the presence of relevant snippets in the clinical notes. In this case, the relevant snippets are “intubation” and “endotracheal intubation”, if the human clinician finds the relevant snippets, he/she will give a positive prediction of code 96.04.

SWAM refers to the same idea of manual coding. As shown in Figure \ref{illustrate}, the first step, through the convolutional layer the model will extract informative snippets that could be relevant to any code. In the second step, the attention layer will assign importance weight to snippets to select the relevant snippets of each code, and in the final step the model summary the weighted score of all relevant snippets of each code to give the predictions of the presence of each code.

\begin{figure*}

 \center
    
  \includegraphics[width=0.95\textwidth]{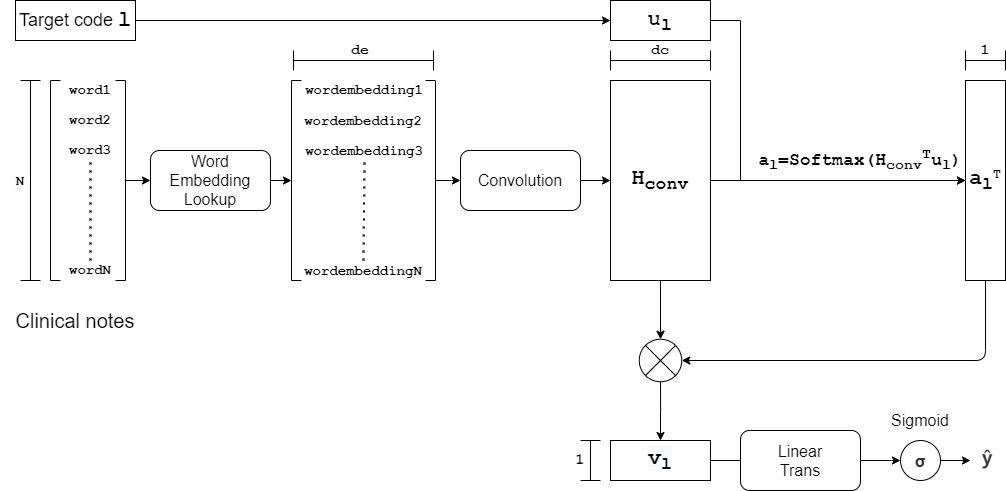}

  \caption{The architecture of the model with per-label mechanism.}

  \label{model}

\end{figure*}

\subsection*{The correspondence between “informative snippet” and convolution filter}
\label{subsection:corres}
Our explanation of the internal mechanics of SWAM builds on the correspondence between “informative snippet” and convolution filter.
Firstly, we classify the "informative snippet" into two categories: "generic snippet" and "non-generic snippet". "generic snippet" refers to snippets that are informative for multiple labels, for example, in our task, "experience fever" is likely to be a "generic snippet" since it is the symptom correlated with multiple diagnoses. "non-generic snippet" refers to snippets that are only considered as informative to a specific or a few labels, for example, in top-50 code task, “endotracheal intubation” will be considered as a "non-generic snippet" since it brings little information gain to the other 49 labels than it brings to the code 96.04 “Insertion of endotracheal tube”.

Then we infer that there exists a correspondence between “informative snippet” and convolution filter, which means one convolution filter can only generate a high activation value for a specific “informative snippet”. Given that in the CNN context, the “informative snippet” can be considered as a set of word embedding sequences that are close in the embedding space. For example, "large mucus plug" and "big mucus plug" are the same "informative snippet" since they have similar meanings and therefore are close in the embedding space. It is most likely that for different "informative snippets", they will have very little chance to be close in the embedding space. For each filter, it will be "highly activated"\\ouput exceeds threshold when the snippet in its window is close to its parameters in embedding space, and this snippet can be considered as the "informative snippet" corresponds to this filter. 

Based on the above inference, an obvious conclusion is that the choice of the width of the convolution layers, a.k.a. the number of convolution filters should depend on the total numbers of "informative snippets" in the task, more accurately, the numbers of "non-generic snippet" since it will much larger than the numbers of "generic snippet" in the large-scale coding task. Besides, empirical guidance in our architecture design is that there could be multiple "non-generic snippets" for each code \cite{gong2018does}.

Therefore we develop Shallow and Wide Attention CNN for this task: the presence of the informative snippet of each code could be considered as a local, low-level feature learned by the shallow CNN, and we also need the convolutional layer to be wide since the model needs to learn the "non-generic snippets" of all codes.

The mechanism behind Shallow and Wide Attention CNN is general for a set of similar text classification tasks that informative snippets relevant to each label scattered at random locations in the input document. So SWAM can be regarded as a general architecture with the following three characteristics, and implement details can be varied (e.g. the attention layer in the model can be either per-label attention mechanism \cite{mullenbach2018expICD} or the full connected layer in textCNN \cite{kim2014convolutional}).

1. The convolutional layer should be sufficiently wide, a.k.a. enough convolution filters to not only extract all generic snippets that are informative for multiple labels, but also all non-generic features that are correlated to specific label and not informative to other labels, the certain number of filters depends on task context, a.k.a. the total number of generic features and non-generic features in the task.

2. The network architecture should be shallow, this model is designed to extract snippets of text, which can be considered as local, low-level features, so a deeper network is unnecessary since informative snippets relevant to each label scattered at random locations in the input document, it is not likely that we can earn any benefit from the global, high-level features by combining the adjacent snippets.

3. Attention mechanism should be introduced to learn the correlations between important/informative snippets and each code.

\subsubsection*{Word embedding}
The word embedding model used in this paper is the word2vec CBOW method by Mikolov et al. \cite{mikolov2013distributed}, we pre-train word embedding of size $d_e=100$ on the preprocessed text from all discharge summaries in MIMIC3, which is the same dataset for training our model. Details about the dataset can be found in \nameref{subsection:dataset}.
We treat ICD code prediction as a multilabel text classification problem \cite{mccallum1999multi}. For clinical note instance i, we want to determine $y_{i,\ell}\in\ {0,\ 1}$ for all $\ell\in\mathcal{L}$. We train a neural network which passes text through a convolutional layer to compute a base representation of the text of each document \cite{kim2014convolutional}, and makes $\ |\mathcal{L}|$ binary classification decisions.

\subsubsection*{Convolutional Layer}
The input of convolutional layer is the clinical notes in form of pre-trained embeddings representing by the matrix $\mathbf{X}=[\mathbf{x}_1,\mathbf{x}_2,\ldots,\mathbf{x}_N]$. The convolution of adjacent embeddings are computed with a convolution filter $W_c\in\mathbb{R}^{k\times d_e\times d_c}$. At step n, we compute
$$\mathbf{h}_n=g(\mathbf{W}_c\ast\mathbf{x}_{n:n+k-1}+\mathbf{b}_c),$$
The input is padded on both sides with zeros so the base representation $\mathbf{H}$ keeps the same length as $\mathbf{X}$. 

\subsubsection*{Attention layer}
\label{attention}
Nowadays attention mechanism has been generalized and has been employed in many different forms \cite{vaswani2017attention}. The core idea of the attention mechanism can be regarded as "giving weight to different parts of the input, to select the part in the input that is more important for the current task". So the full connected layer in textCNN \cite{kim2014convolutional} can also be regarded as a kind of "attention" since 
it weighs input separately for each label.

As we mention in \nameref{subsection:corres}, SWAM can be regarded as a general CNN architecture, and implement details can be varied. We adopt two different implements of the attention layer in our model for different consider considerations. The first one is the per-label attention mechanism by Mullenbach et al. \cite{mullenbach2018expICD}, we adapt it because it can extract snippets from the clinical text as explanations of the model prediction, which can be used to verify our conjecture about the correspondence between “informative snippet” and convolution filter. The second one is the common full connected layer in textCNN \cite{kim2014convolutional}, we adapt it since the textCNN is the basis of many works so it can prove the versatility of SWAM.

For the per-label attention mechanism (the implement shown in Figure \ref{model}), the idea is to calculate the per-label representation of the document and use an attention vector $\mathbf{\alpha}_\ell$ to represents importance distribution over locations in the document.
To obtain the per-label representation of the document, formally a vector parameter $\mathbf{u}_\ell\in\mathbb{R}^{d_c}$ is used to compute the matrix-vector product, $\mathbf{H}^\top\mathbf{u}_\ell$, which can be taken as that the base representation $\mathbf{H}$ is weighted for label $\ell$. The resulting vector is then normalized using a SoftMax operation, obtaining $\mathbf{\alpha}_\ell$, the attention vector, the value of the element in the attention vector is the weighted sum of convolutional features from all kernels in the same place of the document.

$$\mathbf{\alpha}_\ell=\operatorname{SoftMax}(\mathbf{H}^\top\mathbf{u}_\ell)$$

$\mathbf{\alpha}_\ell$ is also taken as the location indicator of the most important snippet for label $\ell$, every element in $\mathbf{\alpha}_\ell$  is corresponding to a location in the document, the value of the element is seen as the importance of the corresponding location for label $\ell$. The highest element value in $\mathbf{\alpha}_\ell$ means the snippet in this location is most important (a.k.a. most informative) for the prediction of label$\ell$. Therefore we obtain an explanation of the prediction in the form of extracted snippets from the document.

$\alpha_{\ell,n}\mathbf{h}_n$, the element-wise vector product is then computed, applies the attention vectors on the base representation to get the vector document representations $\mathbf{v}_\ell$ for label $\ell$,

$$\mathbf{v}_\ell=\sum_{n=1}^{N}{\alpha_{\ell,n}\mathbf{h}_n}$$

For full connected attention, we instead use max-pooling to filter the base representation down to a vector $\mathbf{v}\in d_c$ where every element in $\mathbf{v}$ corresponds to the highest action value of a convolution filter in the text,

$$v _ { j } = \max _ { n } h _ { n , j }$$
\subsubsection*{Classification}
Given the vector representation $v_\ell$, the likelihood for label $\ell$ is computed using a linear layer and a non-linear function sigmoid:
$$\hat { y } _ { \ell } = \sigma \left( \boldsymbol { \beta } _ { \ell } ^ { \top } \boldsymbol { v } _ { \ell } + b _ { \ell } \right)$$

\subsubsection*{Loss function}
The training procedure use BCE (binary cross-entropy) as the loss function, the optimization goal is to minimize the loss.
$$L _ { \mathrm { BCE } } ( X , y ) = - \sum _ { \ell = 1 } ^ { \mathcal { L } } y _ { \ell } \log \left( \hat { y } _ { \ell } \right) + \left( 1 - y _ { \ell } \right) \log \left( 1 - \hat { y } _ { \ell } \right)$$

\subsection*{Why Shallow and Wide Attention CNN}
SWAM can learn a large scale of local and low-level features, it is suitable for the multi-label text classification task that informative snippets relevant to each label are not shared. Also, SWAM addresses the challenges of interpretability: it provides a satisfactory explanation of the internal mechanics of the deep learning method by establishing the correspondence between “informative snippet” and convolution filter.

\section*{Results}
\subsection*{Dataset}
\label{subsection:dataset}
MIMIC-III \cite{johnson2016} is an open-access dataset comprising health data in the form of text and structured records of ICU admissions. Since MIMIC was built, it has become the basis of many works on multi-label classification \cite{wang2016,rajkomar2018scalable}. Following previous works, we train our model on discharge summaries in MIMIC, which summary records about one stay into a single document. We focus on the raw text of the data and ignore the attached features like admission time. Every discharge summary is corresponding to an admission, and each admission is annotated with a set of ICD-9 codes, describing both diagnoses and treatments that occurred during the patient’s stay. We train and evaluate SWAM on a label set consisting of the 50 most frequent labels. We filter the dataset down to the instances that have at least one of the top 50 most frequent codes, after the filter, there are 8,067 summaries for training, 1,574 for validation, and 1,730 for testing. Other detailed statistics for the setting are summarized in Table \ref{tab:desStats}.

\begin{table}[h!]
\caption{Descriptive statistics for MIMIC3}
    \label{tab:desStats}
      \begin{tabular}{ccc}
        \hline
                    & MIMIC-III full  &MIMIC-III 50 \\ \hline
        Training Documents       & 47,724 & 8,067 \\
        Vocabulary Size          & 51,917 & 51,917 \\
        Mean Tokens per Document & 1,485  & 1,530  \\
        Mean Labels per Document & 15.9  & 5.7 \\
        Total Labels             & 8922  & 50   \\
        \hline
      \end{tabular}
\end{table}

\subsubsection*{Preprocessing}

We remove the tokens that contain no alphabetic characters (e.g., removing ‘100’ but keeping ‘100ml’ ). For those tokens that appear too few times to make their semantics difficult to learn, a threshold that only remains tokens that appear in no fewer than 3 training documents is setting, and all tokens that failed the threshold are replaced with an ‘UNK’ token. The distribution of discharge summaries conforms to the long-tailed distribution, 90\% of discharge summaries are short than 2500 tokens, so we truncated discharge summaries to a maximum length of 2500 tokens.

\begin{table*}[t]
\centering
\caption{ Results on MIMIC-III, 50 labels.}
    \label{mainresult}
\begin{tabular}{p{0.14\textwidth}>{\centering}p{0.08\textwidth}>{\centering}p{0.08\textwidth}>{\centering}p{0.08\textwidth}>{\centering}p{0.08\textwidth}>{\centering\arraybackslash}p{0.08\textwidth}}
\hline
\multirow{2}{*}{Model}&\multicolumn{2}{c}{AUC}&\multicolumn{2}{c}{F1}&\multirow{2}{*}{P@5}\\\cline{2-3} \cline{4-5}
&Macro&Micro&Macro&Micro&\\
\hline
C-MemNN \cite{prakash2017condensed}&0.833&-&-&-&0.42\\
Shi et al.(2017) \cite{shi2017towards}&-&0.900&-&0.532&-\\
CAML \cite{mullenbach2018expICD}&0.875&0.909&0.532&0.614&0.609\\
Logistic regression&0.828&0.862&0.477&0.530&0.545\\
SWAM-CAML&\textbf{ 0.900*}&\textbf{ 0.924*}&0.593& 0.648&\textbf{ 0.625*}\\
SWAM-textCNN&0.892&0.919&\textbf{ 0.603*}&\textbf{ 0.652*}&0.620\\
\hline
\end{tabular}
\end{table*}
\subsection*{Baselines}

As mentioned in \nameref{subsection:corres}, SWAM can be regarded as a general CNN architecture and implement details can be varied. In model part \nameref{attention} two different implements of attention layer are adapt for different considerations, we name those two implements as "SWAM-textCNN"\cite{kim2014convolutional} and "SWAM-CAML" \cite{mullenbach2018expICD} separately to indicate the attention approaches they refers.

The baseline we compare against is a bag-of-words logistic regression model, we also compare SWAM-CAML with the origin implement of CAML \cite{mullenbach2018expICD} at the same setting.

For SWAM-textCNN and SWAM-CAML we initialize the embedding weights using the same pre-trained word2vec vectors. The logistic regression model consists of $\ |\mathcal{L}|$ binary one-vs-rest classifiers acting on unigram bag-of-words features. 

\subsubsection*{Parameter tuning}
We tune the hyper-parameters of our SWAM-models using grid search.We sample parameter values for the learning rate $\eta$ ,as well as filter size $k$ , number of filters $ d _ { c }$, and dropout probability $q$ as shown in Table \ref{tab:parameter}. We also adapt hyper-parameter tuning in the previous works as empirical guidance \cite{kim2014convolutional,mullenbach2018expICD,aghaebrahimian2019hyperparameter}. 
We use a fixed batch size of 16, and train the model with early stopping, in the case that the f1-macro does not improve for 10 epochs the training will terminate. 

\begin{table}[h!]
\caption{Hyper-parameter tuning ranges and optimal values for SWAM model}
    \label{tab:parameter}
      \begin{tabular}{ccc}
        \hline
                    & Range  & Optimal Value \\ \hline
        $\eta$       & 0.0001,0.0003,& 0.001 \\
        (learning rate) & 0.001,0.003   &    \\
        $k$          & 1-10          & 4     \\
        (filter size)                        \\
        $d_{c}$      & 50-500        & 500   \\
        (number of filters)                  \\
        $q$          & 0.2-0.8       & 0.2   \\
        (dropout probability)                \\
        \hline
        
      \end{tabular}
\end{table}

\subsection*{Evaluation Metrics}

We focus on two metrics: Macro-averaged F1 and precision at n (denoted as 'P@n'), which is the fraction of the $n$ highest-scored labels that are present in the ground truth. The reason we focus on Macro-averaged F1 is that it pays attention to per-label performance, which can reflect the average performance of the model on different label tasks. As for P@n, we choose it because it reflects the performance of the model as a practical decision support system which presents a fixed number of predicted codes to help user annotated the clinical text. To facilitate comparison with both future and prior work, we also report a variety of metrics includes the area under the ROC curve (AUC) and micro-averaged F1. For recall, Macro-averaged values are calculated by averaging metrics computed per-label. Micro-averaged values are calculated by treating each (document, code) pair as a separate prediction.

\subsection*{Results on quantitative evaluation}

Our main quantitative evaluation involves predicting the 50-code set of ICD-9 codes based on the text of the MIMIC-III discharge summaries. These results are shown in Table \ref{mainresult}. We adopt two different implements of attention layer, named "SWAM-textCNN" and "SWAM-CAML" The SWAM models give the strongest results on all metrics, especially on F1-Macro, which emphasis average performance over different labels. We attribute this improvement to SWAM, by which the wide architecture gives the model ability to more extensively learn the unique features of different codes.

\subsection*{Ablation experiment on width of network}
\begin{figure*}[ht]

 \center

  \includegraphics[width=\textwidth]{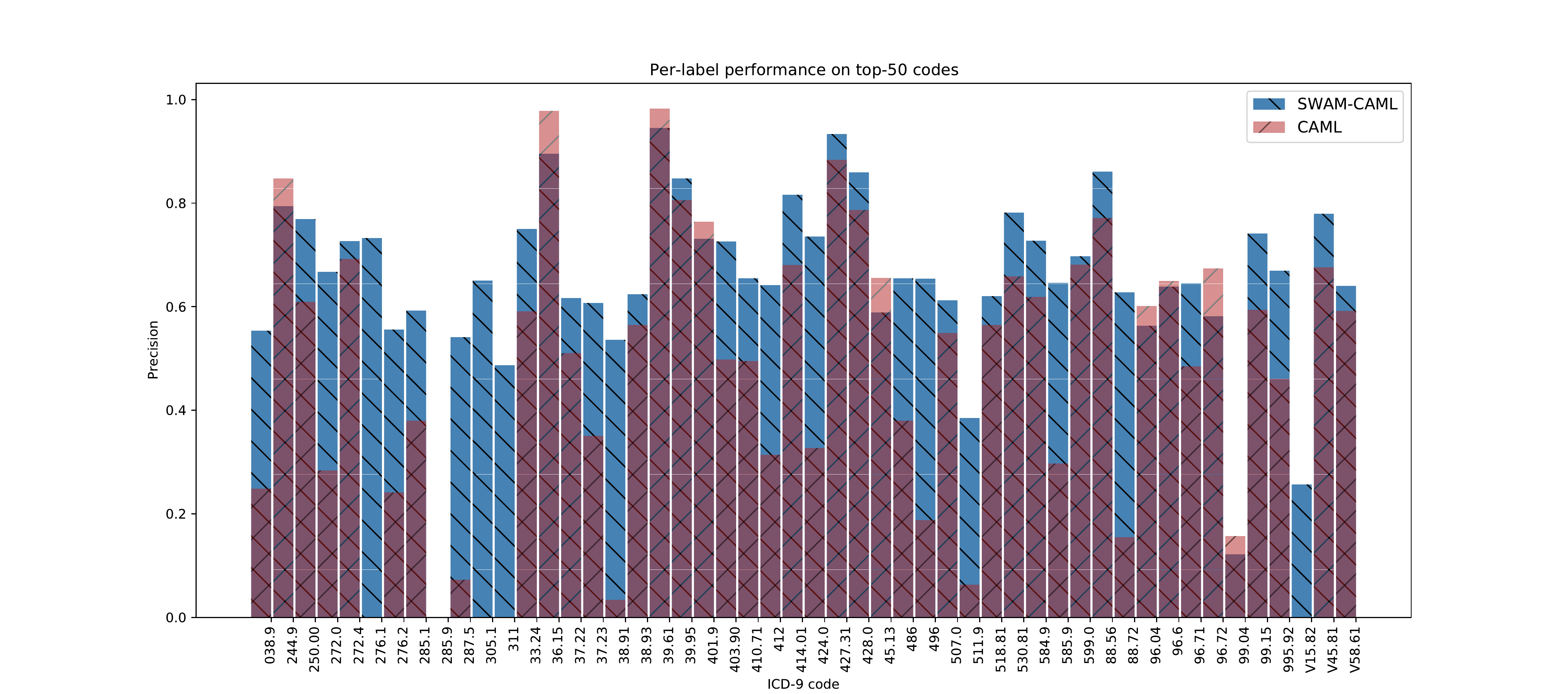}

  \caption{Comparing per-label performances of SWAM-CAML with CAML \cite{mullenbach2018expICD},the only difference between two models are the width of the network. SWAM has 500 filters, while CAML has 50 filters.}

  \label{perlabel}

\end{figure*}
According to our inference about the correspondence between the informative snippet and convolution filter, since each "non-generic snippet" has to correspond to a convolution filter, if the network is too narrow, the model will fail to learn the "non-generic snippets" of some labels. Therefore the impact of the width of the network can be observed from the perspective of per-label performance.

We carry out an ablation experiment on the width of the network, we consider the original implement of CAML \cite{mullenbach2018expICD} as a narrow variant of SWAM, the only difference between SWAM-CAML and CAML are the width of the network, the former has 500 convolution filters and the latter has 50. The experiment results are in line with our expectations. For the narrow model(CAML), the performance of 5 labels is 0, while on the opposite, the wide model(SWAM-CAML) make significant performance improvement that 4 of 5 labels that with a 0 precision in the narrow model now have an average precision of 0.53, at the same time the overall performance of the model is improved. As for the only ICD-9 code 285.9 "Anemia, unspecified" that has 0 precision in both models, we make a manual analysis in Section \nameref{analysis}.

\subsection*{Secondary evaluation}

\subsubsection*{Comparing informative snippets extracted by narrow and wide models}
To verify our inference about \nameref{subsection:corres}, we also compare the informative snippets extracted by both the narrow (CAML) and the wide (SWAM-CAML) model. In order to make the cases representative, from the five labels that has 0 precision in the narrow model, we pick up a label (276.1: Hyposmolality and/or hyponatremia) that has a improved precision in the wide model, and the only label(285.9 Anemia, unspecified) that has 0 precision in both models for analyse.

Table \ref{tab:snippetSample} shows the informative snippets extracted by the SWAM-CAML and the CAML model during the prediction of code 276.1 in two random selected documents. Through a simple analysis, it can be found that the word "hyponatremia" extracted by the SWAM model that appears in both the document and the code description plays an important role in the prediction. While on the opposite, the snippet extracted by the CAML model is not informative since it has 0 precision on this code. 

The word "hyponatremia", as a local and low-level feature can be learned by a single convolution filter according to our inference \nameref{subsection:corres}. Since the only difference between the SWAM-CAML model and the CAML model is the number of filters in the convolutional layer. Table \ref{tab:snippetSample} proves that the performance difference between the narrow and the wide model comes from the learning of non-generic "informative snippet". 

\begin{table}[h!]
    \caption{informative snippets extracted by the wide (SWAM-CAML) and narrow (CAML) models for prediction of ICD code 276.1}
    \label{tab:snippetSample}
       ICD code 276.1: "Hyposmolality and/or hyponatremia"
      \begin{tabular}{p{2cm}p{5cm}}
        \hline
        SWAM-CAML  & …\textbf{dehydration}  and increased abd… \\
        SWAM-CAML  & …\textbf{hyponatremia} and possible initiation of chemotherapy… \\ \hline
        CAML  & …peritonitis renal failure and ileus on the floor the patient was followed by… \\
        CAML  & …renal failure and small bowel obstruction of note the provided information on… \\
        \hline
      \end{tabular}
      
\end{table}

\begin{table}[h!]
    \caption{ICD Code with 0 precision in both  the wide (SWAM-CAML) and narrow (CAML) models before and after data shuffle}
    \label{tab:shuffle}
      \begin{tabular}{p{2cm}p{5.5cm}}
      \hline
        Before shuffle & ICD Code with 0 precision \\
        \hline
        SWAM-CAML  &  285.9 "Anemia, unspecified"\\\hline
        CAML  & 285.9  "Anemia, unspecified" \\
              & V15.82 "History of tobacco use" \\
              & 276.1: "Hyposmolality and/or hyponatremia" \\
              & 305.1  "Tobacco use disorder" \\
              & 311    "Depressive disorder, not elsewhere classified" \\
              
              \hline
        After shuffle & ICD Code with 0 precision\\
        \hline
        SWAM-CAML  &  285.9 "Anemia, unspecified"\\\hline
        CAML  & 285.9  "Anemia, unspecified" \\
              & V15.82 "History of tobacco use" \\
              & \textbf{272.0  "Pure hypercholesterolemia"} \\
              & \textbf{V45.81 "Postsurgical aortocoronary bypass status"} \\
              
        \hline
      \end{tabular}
      
\end{table}

\subsubsection*{Factors determine which "non-generic snippet" will fail to be learned by narrow model}
According to our inference, a narrow network will fail to learn the "non-generic snippet" of a part of labels, which naturally raises a question: what factors determine which "non-generic snippet" will not be learned?
The essence of failing to learn different "non-generic snippet" is that the model converges to different local optimal parameters. The convergence result of the model is related to the distribution of data during training, in other words, the order that "non-generic snippets" appear during training: once a filter learns a specific "non-generic snippet" for some labels, the loss function will encourage it to keep its parameters unchanged, and after all the filters have learned corresponding "informative snippet", it's difficult for the model to leave the current local optimal solution and learn new "informative snippet".Therefore we shuffle the training dataset with a different random seed and re-train the models with the shuffled dataset. The results are shown in Table \ref{tab:shuffle}.

The results in Table \ref{tab:shuffle} are in line with our expectations, after shuffle and re-training, the 0 precision labels in the narrow model (CAML) change. On the opposite, the only 0 precision label 285.9 in the wide model (SWAM-CAML) still can not be learned. The distribution of data during training is a factor that determines which "non-generic snippet" will fail to be learned by the narrow model. And the reason behind the bad performance ICD code 285.9 "Anemia, unspecified" is something beyond the local optimum.

\subsubsection*{Analysis of the reason behind bad performance code}
\label{analysis}

The ICD-9 code 285.9 "Anemia, unspecified" is failed to be predicted by both narrow (CAML) and wide (SWAM) models. Through manual analyzing, we found there are more than 50 codes in the ICD-9 that are in form of “Anemia + specific reason”, which means the snippets related to anemia cannot are necessary but not sufficient for prediction of code 285.9. The prediction of 285.9 Anemia, unspecified is not only based on the presence of snippet related to ‘Anemia’, it is also based on the information that all possible reasons are absent. This is a blind spot in all current machine learning models. It is difficult for models to learn inferences based on missing information.

\section*{Discussion}
For future work, we are considering several different directions.
From the application perspective, since our approach does work well on 50 labels task, the next step is to apply the approach to the full code set. A major challenge is for full code set, we may need tens of thousands of convolution filters, as the number of filters in the network increases, unnecessary overlap in the features captured by the network’s filters will also increase \cite{prakash2019repr}. We plan to address this challenge by adapting the ensemble method, we plan to cluster the ICD codes and train a classifier for each clustered subset.
From the linguistic perspective, we plan to explore reasons behind hard-to-learn codes such as ICD-9 code 285.9 "Anemia, unspecified", and leverage the hierarchy of ICD codes to improve performance on these codes.

\section*{Conclusion}
We present SWAM, an explainable CNN approach for multi-label document classification, which employs a wide convolution layer to learn local and low-level features for each label, yields strong improvements over previous metrics on the ICD-9 code prediction task, while providing satisfactory explanations for its internal mechanics.

\nocite{}


\begin{backmatter}

\section*{Abbreviations}
ICD:International Classification of Diseases;CNN:Convolutional Neural Network;ICU:Intensive care unit;SWAM:Shallow and Wide Attention convolutional Mechanism;MIMIC-III:Medical Information Mart for Intensive Care III;NLP:Natural Language;CBOW:Continuous Bag-of-Words;BCE:Binary cross-entropy;CAML:Convolutional Attention for Multi-Label classification;F1:F-measure

\section*{Availability of data and materials}
The datasets is available from \url{https://mimic.physionet.org/}.

\section*{Competing interests}
The authors declare that they have no competing interests.

\section*{Author's contributions}
Hu S.Y. conceived of the presented idea, developed the theory and carry out the experiments. Teng F. encouraged Hu S.Y. to investigate per-label performance of the model and supervised the findings of this work. All authors discussed the results. Hu S.Y. wrote the manuscript under supervising from Teng F., all authors have read and approved the final manuscript.

\section*{Funding}
No funding to declare. 

\section*{Ethics approval and consent to participate}
Not applicable.

\section*{Consent for publication}
Not applicable.

\section*{Acknowledgements}
None. 

\bibliographystyle{vancouver} 
\bibliography{bmc_article}      

\end{backmatter}
\end{document}